\documentclass{article}
\PassOptionsToPackage{numbers,compress}{natbib}
\usepackage[preprint]{nips-template/neurips_2026}

\usepackage[utf8]{inputenc}
\usepackage[T1]{fontenc}
\usepackage{microtype}
\usepackage{hyperref}
\usepackage{url}
\usepackage{booktabs}
\usepackage{amsmath}
\usepackage{amsfonts}
\usepackage{multirow}
\usepackage{graphicx}
\usepackage{subfigure}
\usepackage{xcolor}
\usepackage{eso-pic}
\definecolor{darkblue}{rgb}{0, 0, 0.5}
\hypersetup{colorlinks=true, citecolor=darkblue, linkcolor=darkblue, urlcolor=darkblue}

\title{Balanced Aggregation: Understanding and Fixing Aggregation Bias in GRPO}

\author{%
\textbf{Zhiyuan Zeng}$^{1,3}$ \quad
\textbf{Jiameng Huang}$^{2}$ \quad
\textbf{Zhangyue Yin}$^{1}$ \quad
\textbf{Jiashuo Liu}$^{3}$ \quad
\textbf{Ziniu Li}$^{3}$\\
\textbf{Bingrui Li}$^{4}$ \quad
\textbf{Yuhao Wu}$^{6}$ \quad
\textbf{Yining Zheng}$^{1}$ \quad
\textbf{Ge Zhang}$^{3}$\thanks{Corresponding authors: Ge Zhang, Wenhao Huang, and Xipeng Qiu.} \quad
 \textbf{Wenhao Huang}$^{3}$\footnotemark[1] \quad
 \textbf{Xipeng Qiu}$^{1,5}$\footnotemark[1] \\
\parbox{0.95\linewidth}{\centering
$^{1}$Fudan University \quad
$^{2}$Peking University \quad
$^{3}$M-A-P \quad
$^{4}$Tsinghua University \\
$^{5}$Shanghai Innovation Institute \quad
$^{6}$Singapore University of Technology and Design
}\\
\small
\href{mailto:cengzy23@m.fudan.edu.cn}{\texttt{cengzy23@m.fudan.edu.cn}} \quad
\href{mailto:gezhang@umich.edu}{\texttt{gezhang@umich.edu}} \quad
\href{mailto:rubio8741@gmail.com}{\texttt{rubio8741@gmail.com}} \quad
\href{mailto:xpqiu@fudan.edu.cn}{\texttt{xpqiu@fudan.edu.cn}}
}

\begin{document}
\AddToShipoutPictureBG*{%
  % --- 左上角的 Logo ---
  \AtPageUpperLeft{%
    \hspace{1.5cm}%
    \raisebox{-3cm}{%
      \includegraphics[width=2cm]{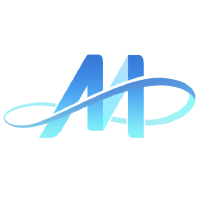}%
    }%
  }
  % --- 模拟右上角的 Logo ---
  \AtPageUpperLeft{%
    % \paperwidth 是整个纸张的宽度
    \hspace{\paperwidth}% 先移动到最右边
    \hspace{-3cm}% 再向左移动，留出边距
    \raisebox{-2.5cm}{%
      % 使用 \rlap 来确保图片向左绘制，不占据右侧空间
      \rlap{\includegraphics[width=1cm]{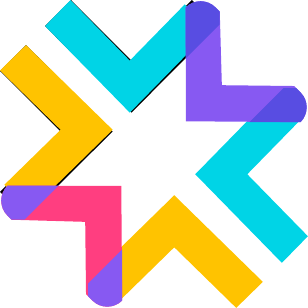}}%
    }%
  }
}

\maketitle

\begin{abstract}
Reinforcement learning with verifiable rewards (RLVR) has become a central paradigm for improving reasoning and code generation in large language models, and GRPO-style training is widely adopted for its simplicity and effectiveness. However, an important design choice remains underexplored: how token-level policy gradient terms are aggregated within each sampled group. Standard GRPO uses sequence aggregation, while recent work has advocated token aggregation as a better alternative. We show that these two rules induce different optimization biases: token aggregation introduces sign-length coupling, while sequence aggregation implicitly downweights longer responses through sequence-level equal weighting. To address this tension, we propose \textbf{Balanced Aggregation (BA)}, a simple drop-in replacement that computes token-level means separately within the positive and negative subsets and then combines them with sequence-count-based weights. Experiments with Qwen2.5-Math-7B and Qwen3-1.7B on DAPO-17k and Polaris, evaluated on six reasoning and coding benchmarks, show that BA consistently improves training stability and final performance over standard token and sequence aggregation. Our analysis further shows that the relative effectiveness of token and sequence aggregation is largely governed by response-length variation and the positive-negative length gap, highlighting aggregation as a critical design dimension in GRPO-style RLVR.
\end{abstract}

\section{Introduction}
Reinforcement learning with verifiable rewards (RLVR) has become a central paradigm for improving the reasoning and code generation abilities of large language models (LLMs).
By replacing learned reward models with programmatically verifiable signals such as exact-match correctness or unit-test pass rate, RLVR provides a simple and scalable way to optimize models on tasks with objective outcomes.

Among recent RLVR methods, GRPO-style training is particularly attractive in practice due to its simplicity and effectiveness.
For each prompt, the policy samples multiple responses, assigns rewards based on verifiable outcomes, and computes normalized group-wise advantages to optimize a PPO-style objective.
This design has been widely adopted in reasoning and coding settings because it avoids training a separate critic while still providing useful relative learning signals within each sampled group.

Despite the growing adoption of GRPO-style RLVR, an important design choice remains underexplored: \emph{how token-level policy gradient terms are aggregated within each sampled group}. In standard GRPO, the default choice is \emph{sequence aggregation}, which first averages over tokens within each response and then averages across responses. Recent works such as DAPO and Dr.GRPO highlighted limitations of this design and accordingly advocated \emph{token aggregation}, which directly averages the clipped objective over all tokens in the sampled group, as a better alternative \citep{yu2025dapoopensourcellmreinforcement, liu2025understandingr1zeroliketrainingcritical}. In this paper, we show that these two rules induce systematically different optimization biases and can lead to substantially different training dynamics and final performance.

We show that token aggregation introduces a \emph{sign-length coupling bias}: the relative contribution of positive and negative samples on the policy gradient depends not only on their normalized advantages, but also on their average response lengths.
Therefore, when positive and negative responses have different length distributions, token aggregation can systematically amplify one side of the update.

Sequence aggregation removes this positive-negative length coupling by assigning equal weight to each response.
However, this introduces a different bias: longer responses are implicitly downweighted because each sequence contributes equally regardless of how many tokens it contains \citep{yu2025dapoopensourcellmreinforcement, liu2025understandingr1zeroliketrainingcritical}.

These two biases matter in practice.
We find that token aggregation can be favorable when response length variance is large, since it avoids overly suppressing long responses.
However, it is also more sensitive to positive-negative length imbalance and often leads to less stable optimization.
This tension suggests that a better aggregation rule should preserve the sign-balance property of sequence aggregation without inheriting its strong sequence-level equal-weighting effect.

To this end, we propose \textbf{Balanced Aggregation (BA)}.
The key idea is simple: we first split responses within each group into positive and negative subsets according to the sign of their normalized advantages, compute token-level means separately within each subset, and then combine the two subset losses using weights proportional to the number of sequences in each subset.
This construction removes the positive-negative length coupling induced by token aggregation, while retaining token-level averaging within each sign group.
As a result, BA preserves the same inter-sign balancing principle as sequence aggregation, but does not force every response to have equal weight within a sign group.

We evaluate BA on GRPO-style RLVR training using Qwen2.5-Math-7B and Qwen3-1.7B across DAPO and Polaris training sets, and report results on six evaluation benchmarks including Math-500, AIME 2024, AIME 2025, OlympicBench, Minerva-MATH, and LiveCodeBench.
Across both weak and strong model regimes, BA consistently delivers stronger final performance and better training stability than standard token and sequence aggregation.
Our analysis further shows that the relative effectiveness of token and sequence aggregation can be largely explained by two factors: response length variance and the response-length gap between positive and negative samples.

Our contributions are as follows:

\begin{itemize}
    \item We show that loss aggregation in GRPO-style RLVR is not a benign implementation detail, and provide a unified analysis of the sign-length coupling bias in token aggregation and the sequence equal-weighting bias in sequence aggregation.
    \item We propose Balanced Aggregation (BA), a simple drop-in replacement that performs token-level averaging separately within the positive and negative subsets before combining them, thereby avoiding the main bias of token aggregation without imposing the strong equal-weighting effect of sequence aggregation.
    \item We provide extensive empirical evidence that BA improves robustness and final performance across models, datasets, and evaluation benchmarks, and clarify when token aggregation or sequence aggregation is preferable.
\end{itemize}

\section{Related Work}

\paragraph{RLVR and GRPO-Style Post-Training}
Recent progress in reasoning-oriented LLM post-training has highlighted the importance of reinforcement learning, as reflected by the success of systems such as OpenAI's o1, DeepSeek-R1 \citep{openai2024openaio1card,Guo_2025,zeng2024scalingsearchlearningroadmap}. In tasks with programmatically verifiable outcomes, reinforcement learning with verifiable rewards (RLVR) has emerged as a particularly attractive paradigm because it avoids learned reward modeling and provides a scalable training signal for reasoning and code generation. On the optimization side, PPO~\citep{schulman2017proximal} has long served as the standard policy optimization backbone, while GRPO, introduced in DeepSeekMath~\citep{shao2024deepseekmath}, further reduces training cost by replacing the critic with group-relative reward normalization. This critic-free formulation has made GRPO-style training a practical foundation for large-scale RLVR.

\paragraph{RLVR Training Tricks}
A growing line of work has improved RLVR training from multiple angles. Some methods focus on reducing train-infer mismatch and improving training stability \citep{yao2025offpolicy,liu-li-2025-rl-collapse,ma2025stabilizingmoereinforcementlearning,deepseekai2025deepseekv32pushingfrontieropen,zhiyuan2025rloopselfimprovingframeworkreinforcement}. Another line studies clipping and trust-region design, including asymmetric clipping \citep{yu2025dapoopensourcellmreinforcement} and soft clipping \citep{minimax2025minimaxm1scalingtesttimecompute}. A related direction directly improves importance sampling, with methods such as GSPO and ASPO \citep{zheng2025groupsequencepolicyoptimization,wang2025aspoasymmetricimportancesampling}. Some works also examine the role of advantage estimation and normalization, showing that the standard-deviation normalization can introduce nontrivial optimization bias \citep{liu2025understandingr1zeroliketrainingcritical,deepseekai2025deepseekv32pushingfrontieropen}. Recent empirical studies further revisit a wide range of RLVR training heuristics and scaling choices, highlighting that many commonly used tricks can have subtle effects \citep{liu2025itrickstrapsdeep,khatri2025artscalingreinforcementlearning}.
Together, these studies show that RLVR performance depends heavily on a collection of low-level optimization choices rather than on the policy objective alone.

\paragraph{Aggregation in GRPO-Style RL}
Among these design choices, how token-level policy gradient terms are aggregated has received less systematic attention. Standard GRPO uses sequence aggregation, which first averages token-level contributions within each response and then averages across responses. Recent works such as DAPO and Dr.GRPO identified limitations of this design in long-form reasoning and accordingly advocated token-level alternatives~\citep{yu2025dapoopensourcellmreinforcement, liu2025understandingr1zeroliketrainingcritical}. GMPO improves optimization stability in a different way, by replacing the arithmetic mean with a geometric mean \citep{zhao2025geometricmeanpolicyoptimization}. By contrast, our focus is the bias induced by the aggregation rule itself, so we center the analysis and experiments on sequence aggregation and token aggregation, and position Balanced Aggregation as a simple alternative that directly addresses their respective biases.

\section{Method}

\subsection{GRPO-style RLVR}

We consider reinforcement learning with verifiable rewards in the standard group-based setting.
Given an input prompt \(x\), the current policy \(\pi_\theta\) samples a group of \(G\) responses:
\begin{equation}
o_1, o_2, \dots, o_G \sim \pi_\theta(\cdot \mid x).
\end{equation}

Each response \(o_i\) receives a scalar reward \(r_i\), computed by a verifiable reward function such as exact-match correctness for math or unit-test pass rate for code.
GRPO normalizes rewards within each group to produce sequence-level advantages.
Let
\begin{equation}
\mu = \frac{1}{G}\sum_{i=1}^G r_i, \qquad
\sigma = \sqrt{\frac{1}{G}\sum_{i=1}^G (r_i - \mu)^2 + \epsilon},
\end{equation}
then the normalized advantage for response \(i\) is
\begin{equation}
\hat A_i = \frac{r_i - \mu}{\sigma}.
\end{equation}

Importantly, \(\hat A_i\) is defined at the \emph{sequence level}, so all tokens in the same response share the same advantage.

\subsection{Token-Level PPO Objective}

Let response \(o_i\) contain \(T_i\) generated tokens.
For token \(t \in \{1,\dots,T_i\}\), define the policy ratio
\begin{equation}
\rho_{i,t}(\theta)
=
\frac{
\pi_\theta(o_{i,t} \mid x, o_{i,<t})
}{
\pi_{\theta_{\mathrm{old}}}(o_{i,t} \mid x, o_{i,<t})
}.
\end{equation}

The token-level clipped PPO contribution is
\begin{equation}
\phi_{i,t}(\theta)
=
\min\!\Big(
\rho_{i,t}(\theta)\hat A_i,\;
\mathrm{clip}(\rho_{i,t}(\theta), 1-\epsilon, 1+\epsilon)\hat A_i
\Big).
\end{equation}

The full GRPO-style objective is obtained by aggregating these token-level terms across the sampled group.

\subsection{Aggregation rules}
\label{sec:aggregation-rules}

The aggregation rule determines how the token-level contributions \(\phi_{i,t}(\theta)\) are combined into a group-level loss.
This choice is especially important because the advantage is sequence-level, while the objective is token-level.
Different aggregation schemes therefore imply different weighting structures over responses and tokens.

A common choice is \textbf{token aggregation}, which averages over all tokens in the group:
\begin{equation}
\mathcal J_{\mathrm{token}}(\theta)
=
\mathbb E\left[
\frac{1}{N}
\sum_{i=1}^G \sum_{t=1}^{T_i} \phi_{i,t}(\theta)
\right],
\qquad
N = \sum_{i=1}^G T_i.
\end{equation}

Another common choice is \textbf{sequence aggregation}, which first averages within each response and then averages across responses:
\begin{equation}
\mathcal J_{\mathrm{seq}}(\theta)
=
\mathbb E\left[
\frac{1}{G}
\sum_{i=1}^G
\frac{1}{T_i}
\sum_{t=1}^{T_i} \phi_{i,t}(\theta)
\right].
\end{equation}

Although both objectives optimize the same token-level PPO term, they correspond to different implicit weighting schemes.
In the following, we formalize this difference and introduce a balanced alternative.

\subsection{Motivation: Aggregation Bias in GRPO}
\label{sec:aggregation-bias}

In GRPO-style RLVR, the normalized advantage \(\hat A_i\) is shared by all tokens in response \(i\), while the PPO objective is computed at the token level.
Therefore, the group-level aggregation rule directly determines how response length affects the relative weight of different samples.

To make this explicit, we partition the sampled group into positive and negative subsets:
\begin{equation}
S_+ = \{ i \mid \hat A_i > 0 \},
\qquad
S_- = \{ i \mid \hat A_i < 0 \}.
\end{equation}
Let
\begin{equation}
k = |S_+|, \qquad G-k = |S_-|.
\end{equation}

For analysis, we write the token-level contribution as
\begin{equation}
\phi_{i,t}(\theta) = \hat A_i \, \delta_{i,t}(\theta),
\end{equation}
where \(\delta_{i,t}(\theta)\) denotes the effective token-level PPO term after factoring out the sequence-level advantage.
Under the standard binary-reward GRPO setting, the normalized advantages take the form
\begin{equation}
\hat A_i =
\sqrt{\frac{G-k}{k}}
\quad \text{for } i \in S_+,
\qquad
\hat A_i =
-\sqrt{\frac{k}{G-k}}
\quad \text{for } i \in S_-.
\end{equation}

Under token aggregation, the objective can be rearranged as
\begin{equation}
\mathcal J_{\mathrm{token}}
\propto
\frac{\sqrt{k(G-k)}}{N}
\left(
\bar T_+ \, \bar \delta_+^{\mathrm{tok}}
-
\bar T_- \, \bar \delta_-^{\mathrm{tok}}
\right),
\end{equation}
where
\begin{equation}
\bar T_+ = \frac{1}{k}\sum_{i \in S_+} T_i,
\qquad
\bar T_- = \frac{1}{G-k}\sum_{i \in S_-} T_i,
\end{equation}
and
\begin{equation}
\bar \delta_+^{\mathrm{tok}}
=
\frac{1}{N_+}\sum_{i \in S_+}\sum_{t=1}^{T_i}\delta_{i,t},
\qquad
\bar \delta_-^{\mathrm{tok}}
=
\frac{1}{N_-}\sum_{i \in S_-}\sum_{t=1}^{T_i}\delta_{i,t},
\end{equation}
with
\begin{equation}
N_+ = \sum_{i \in S_+} T_i,
\qquad
N_- = \sum_{i \in S_-} T_i.
\end{equation}

This expression reveals a \textbf{sign-length coupling bias}: the positive and negative terms are weighted by \(\bar T_+\) and \(\bar T_-\), so their relative contribution depends on the average response lengths of the two sign groups.
As a result, when \(\bar T_+ \neq \bar T_-\), token aggregation changes the effective balance of policy gradients; in Section~\ref{sec:analysis}, we will show that this bias is reflected in the policy-gradient loss dynamics (Figure~\ref{fig:pg_loss_evidence}).

Sequence aggregation removes this coupling at the positive-negative group level:
\begin{equation}
\mathcal J_{\mathrm{seq}}
\propto
\frac{\sqrt{k(G-k)}}{G}
\left(
\bar \delta_+^{\mathrm{seq}} - \bar \delta_-^{\mathrm{seq}}
\right),
\label{eq:seq-balanced-form}
\end{equation}
where
\begin{equation}
\bar \delta_+^{\mathrm{seq}}
=
\frac{1}{k}
\sum_{i \in S_+}
\left(
\frac{1}{T_i}\sum_{t=1}^{T_i}\delta_{i,t}
\right),
\qquad
\bar \delta_-^{\mathrm{seq}}
=
\frac{1}{G-k}
\sum_{i \in S_-}
\left(
\frac{1}{T_i}\sum_{t=1}^{T_i}\delta_{i,t}
\right).
\label{eq:seq-within-sign}
\end{equation}

Thus, sequence aggregation equalizes the relative weight of positive and negative responses at the group level, but it does so by assigning equal weight to each sequence regardless of its token count.
We refer to this as \textbf{sequence equal-weighting bias}, which is also related to observations made in DAPO \citep{yu2025dapoopensourcellmreinforcement} and Dr.GRPO \citep{liu2025understandingr1zeroliketrainingcritical}.
These observations suggest that neither token aggregation nor sequence aggregation is fully satisfactory: the former couples sign and length, while the latter removes that coupling by imposing strong per-sequence equal weighting. As we show later in Section~\ref{sec:agg-theory-empirical} using Figure~\ref{fig:length_analysis}, the relative impact of these two biases directly shapes RLVR performance across different model regimes.

\subsection{Balanced Aggregation}

We propose \textbf{Balanced Aggregation (BA)}, a simple aggregation rule that separates positive and negative samples before averaging.

We first compute token-level mean losses within the positive and negative subsets:
\begin{equation}
\mathcal L_+
=
\frac{1}{N_+}
\sum_{i \in S_+} \sum_{t=1}^{T_i} \phi_{i,t}(\theta),
\qquad
\mathcal L_-
=
\frac{1}{N_-}
\sum_{i \in S_-} \sum_{t=1}^{T_i} \phi_{i,t}(\theta).
\end{equation}

We then combine them using sequence-count-based weights:
\begin{equation}
\mathcal J_{\mathrm{BA}}(\theta)
=
\mathbb E\left[
\frac{k}{G}\mathcal L_+
+
\frac{G-k}{G}\mathcal L_-
\right].
\end{equation}

Equivalently,
\begin{equation}
\mathcal J_{\mathrm{BA}}(\theta)
=
\mathbb E\left[
\frac{k}{G}
\cdot
\frac{1}{N_+}
\sum_{i \in S_+}\sum_{t=1}^{T_i}\phi_{i,t}(\theta)
+
\frac{G-k}{G}
\cdot
\frac{1}{N_-}
\sum_{i \in S_-}\sum_{t=1}^{T_i}\phi_{i,t}(\theta)
\right].
\end{equation}

The intuition is straightforward.
Within each sign group, BA retains token-level averaging, so it does not force every response to have equal weight.
Across sign groups, BA uses sequence-count-based reweighting, which restores the same positive-negative balancing principle as sequence aggregation. In particular, the weights \(k/G\) and \((G-k)/G\) are chosen so that, under the binary-reward GRPO setting, BA induces the same inter-sign prefactor as sequence aggregation; a short derivation is provided in Appendix~\ref{app:ba-weights}.

\subsection{Connection to Sequence Aggregation}

BA is closely related to sequence aggregation, but the two are not equivalent.

Under the same binary-reward GRPO setting, substituting the normalized advantages into BA yields
\begin{equation}
\mathcal J_{\mathrm{BA}}
\propto
\frac{\sqrt{k(G-k)}}{G}
\left(
\bar \delta_+^{\mathrm{BA}} - \bar \delta_-^{\mathrm{BA}}
\right),
\end{equation}
where
\begin{equation}
\bar \delta_+^{\mathrm{BA}}
=
\frac{1}{N_+}
\sum_{i \in S_+}\sum_{t=1}^{T_i}\delta_{i,t},
\qquad
\bar \delta_-^{\mathrm{BA}}
=
\frac{1}{N_-}
\sum_{i \in S_-}\sum_{t=1}^{T_i}\delta_{i,t}.
\end{equation}

By contrast, sequence aggregation has exactly the same inter-sign form as in Eq.~\eqref{eq:seq-balanced-form}, with within-sign averages defined in Eq.~\eqref{eq:seq-within-sign}.

Therefore, BA and sequence aggregation share the same \emph{inter-sign balancing} structure: both remove the sign-length coupling of token aggregation and induce the same positive-negative prefactor \(\sqrt{k(G-k)}/G\).
However, they differ in their \emph{within-sign averaging} rule.
Sequence aggregation gives equal weight to each response within a sign group, whereas BA averages over all tokens within that sign group.

In general,
\begin{equation}
\bar \delta_{\pm}^{\mathrm{seq}}
\neq
\bar \delta_{\pm}^{\mathrm{BA}},
\end{equation}
unless all responses within a sign group have the same length. Thus, BA should be understood as preserving the sign-balance property of sequence aggregation without inheriting its strong per-sequence equal-weighting effect.

BA is a simple drop-in replacement for the aggregation step in GRPO-style RLVR. It removes the sign-length coupling bias of token aggregation while avoiding the strong sequence equal-weighting bias of sequence aggregation. Although the current formulation of BA is derived under the binary-reward setting, BA can naturally extend to non-binary rewards, which is shown in Appendix~\ref{app:nonbinary}.

\section{Experiments}
\subsection{Experimental Settings}

\paragraph{Training Data}
We conduct RLVR training on two datasets: \textbf{DAPO-17k} (approximately 17,000 mathematical reasoning problems) and \textbf{Polaris} (approximately 53,000 mathematical problems) \citep{yu2025dapoopensourcellmreinforcement,polaris2025recipe}. Both consist of problem‑answer pairs, where answers are used to compute verifiable rewards for generated responses.

\paragraph{Evaluation Benchmarks}
We evaluate on six benchmarks covering both difficult reasoning and coding tasks: Math‑500, AIME‑2024, AIME‑2025, OlympicBench, Minerva‑MATH, and LivecodeBench \citep{lightman2023lets,Lewkowycz2022SolvingQR,He2024OlympiadBenchAC,jain2024livecodebench}.

\paragraph{Compared Methods}
We compare three aggregation rules applied within the DAPO algorithm:
\begin{itemize}
    \item \textbf{token‑agg}: Token‑level averaging, where the clipped PPO objective is averaged over all tokens in the sampled group. This is used in DAPO and Dr.GRPO \citep{yu2025dapoopensourcellmreinforcement,liu2025understandingr1zeroliketrainingcritical}.
    \item \textbf{seq‑agg}: Sequence‑level averaging, where token‑level contributions are first averaged within each response and then averaged across responses. This is the default in GRPO \citep{shao2024deepseekmath}.
    \item \textbf{balanced‑agg}: Our proposed balanced aggregation, which splits responses by advantage sign, computes token‑level means separately within positive and negative subsets, and combines them with sequence‑count‑based weights.
\end{itemize}
All other components (advantage normalization, PPO clipping, sampling) are kept identical across methods.

\paragraph{Training Details}
We train \textbf{Qwen2.5‑Math‑7B} and \textbf{Qwen3‑1.7B} \citep{yang2024qwen25math,yang2025qwen3} with maximum response lengths of 2,048 and 8,192 tokens. Training is implemented in the verl framework, using group size $G = 16$, learning rate $10^{-6}$, and 500 total steps. We apply PPO clipping bounds of 0.2 and 0.28 (the standard DAPO setting). The global batch size is 128 prompts, each generating 16 responses via vLLM (temperature 1.0). Other hyper‑parameters follow the standard DAPO configuration \citep{yu2025dapoopensourcellmreinforcement}.

\paragraph{Evaluation Protocol}
We sample 8 responses per prompt (temperature 1.0). For math benchmarks, correctness is determined using OpenCompass's rule‑based verifier \citep{opencompass2024}; for LivecodeBench, we execute the generated code against unit tests \citep{jain2024livecodebench}. We report three metrics: \textbf{peak accuracy} (highest accuracy observed during training), \textbf{peak best@8 accuracy}, and \textbf{last‑step accuracy} (accuracy at the final training step).

\subsection{Main Results}
\label{sec:main-results}

To evaluate \textbf{Balanced Aggregation} (denoted as \textbf{balanced-agg}), we compare it against \textbf{token-agg} and \textbf{seq-agg} baselines on two training datasets: DAPO-17k and Polaris. We benchmark the aggregation methods on two base models: Qwen2.5-Math-7B and Qwen3-1.7B.

Table~\ref{tab:main-results} presents the average scores across six evaluation benchmarks. Since a full breakdown would necessitate an overly large table, detailed per-benchmark results are shown in Figure \ref{fig:peak_vs_last}. Furthermore, because RLVR training dynamics can be highly volatile in later stages, the highest peak performance does not guarantee the best last-step performance. We therefore explicitly report both \textbf{peak} and \textbf{last-step} metrics to comprehensively evaluate each method's training stability.

\begin{table}[tb]
\centering
\caption{Main experimental results on DAPO-17k and Polaris datasets using Qwen2.5-Math-7B and Qwen3-1.7B models. We report the average peak and last-step accuracy across six evaluation benchmarks.}
\label{tab:main-results}
\small
\setlength{\tabcolsep}{3pt}
\begin{tabular}{llcccc}
\toprule
& & \multicolumn{2}{c}{\textbf{Peak Performance}} & \multicolumn{2}{c}{\textbf{Last-Step}} \\
\cmidrule(lr){3-4} \cmidrule(lr){5-6}
\textbf{Dataset} & \textbf{Model / Aggregation} & \textbf{Acc@8} & \textbf{Best@8} & \textbf{Acc@8} & \textbf{Best@8} \\
\midrule
\multirow{8}{*}{\textbf{DAPO-17k}}
& \textit{Qwen2.5-Math-7B} & & & & \\
& \quad seq-agg          & 0.3578 & 0.4821 & \textbf{0.3446} & 0.4598 \\
& \quad token-agg        & 0.3595 & 0.4869 & 0.3364 & 0.4434 \\
& \quad balanced-agg     & \textbf{0.3634} & \textbf{0.4896} & 0.3424 & \textbf{0.4627} \\
\cmidrule{2-6}
& \textit{Qwen3-1.7B} & & & & \\
& \quad seq-agg               & \textbf{0.4947} & 0.5990 & 0.4481 & 0.5522 \\
& \quad token-agg             & 0.4876 & 0.6058 & 0.4360 & 0.5608 \\
& \quad balanced-agg          & 0.4928 & \textbf{0.6086} & \textbf{0.4695} & \textbf{0.5861} \\
\midrule
\multirow{8}{*}{\textbf{Polaris}}
& \textit{Qwen2.5-Math-7B} & & & & \\
& \quad seq-agg          & 0.3405 & 0.4697 & 0.3172 & 0.4360 \\
& \quad token-agg        & \textbf{0.3539} & \textbf{0.4768} & 0.3292 & \textbf{0.4527} \\
& \quad balanced-agg     & 0.3423 & 0.4750 & \textbf{0.3319} & 0.4443 \\
\cmidrule{2-6}
& \textit{Qwen3-1.7B} & & & & \\
& \quad seq-agg               & 0.4812 & 0.5950 & 0.4614 & \textbf{0.5730} \\
& \quad token-agg             & 0.4842 & 0.6053 & 0.4349 & 0.5557 \\
& \quad balanced-agg          & \textbf{0.4879} & \textbf{0.6094} & \textbf{0.4640} & 0.5550 \\
\bottomrule
\end{tabular}
\end{table}

\paragraph{Overall results on DAPO-17k}
For Qwen2.5-Math-7B, \textbf{token-agg} yields better peak performance than \textbf{seq-agg}, but \textbf{balanced-agg} surpasses both to establish the highest peak metrics. For Qwen3-1.7B, the relationship flips: \textbf{seq-agg} becomes superior to \textbf{token-agg}. This is highly relevant since \textbf{token-agg} is the default in frameworks like \textit{verl}, yet it is clearly not universally better than seq-agg. More crucially, while \textbf{balanced-agg} achieves peak metrics comparable to \textbf{seq-agg}, it successfully prevents the severe degradation often observed in later stages of RLVR, maintaining much higher last-step accuracies.

\paragraph{Overall results on Polaris}
Similar performance dynamics are observed on Polaris. For Qwen2.5-Math-7B, \textbf{token-agg} exhibits the highest peak metrics but suffers noticeable degradation toward the end of training. In contrast, \textbf{balanced-agg} achieves the most robust last-step accuracy and strictly outperforms \textbf{seq-agg} across all evaluated metrics. For Qwen3-1.7B, \textbf{balanced-agg} achieves the highest peak metrics compared to both baselines. In the final training stages, while \textbf{token-agg} suffers a severe collapse, both \textbf{seq-agg} and \textbf{balanced-agg} maintain much more robust last-step accuracies.

\paragraph{Cross-setting summary}
Across both datasets, a consistent dynamic emerges: \textbf{token-agg} performs better on Qwen2.5-Math-7B, whereas \textbf{seq-agg} is more stable and accurate on Qwen3-1.7B. This suggests neither standard aggregation provides a consistently reliable optimization signal across different base models. We delve into the reasons behind this performance flip in Section~\ref{sec:analysis}. 
\textbf{Balanced-agg} successfully bridges this gap, consistently ranking as the best or highly competitive method across our evaluated models and datasets. Its ability to simultaneously preserve within-sign token-level averaging while removing the positive-negative length coupling substantially improves training stability.

\subsection{Analysis}
\label{sec:analysis}

The main results in Table~\ref{tab:main-results} show that aggregation rules interact strongly with the base model and training corpus, and that peak accuracy alone can be misleading when training is volatile.
In this subsection, we unpack these findings along three complementary axes.
First, we compare peak versus last-step accuracy at the per-benchmark level to make training stability visually explicit.
Second, we connect the observed optimization behavior to our theoretical account by examining policy-gradient loss trajectories during training.
Finally, we connect the theory in Section~\ref{sec:aggregation-bias} to the model-dependent flip in Section~\ref{sec:main-results} and to length statistics over training (Figure~\ref{fig:length_analysis}).

\subsubsection{Peak vs. Last-Step Performance}
\begin{figure*}[tb]
\centering
\includegraphics[width=\linewidth]{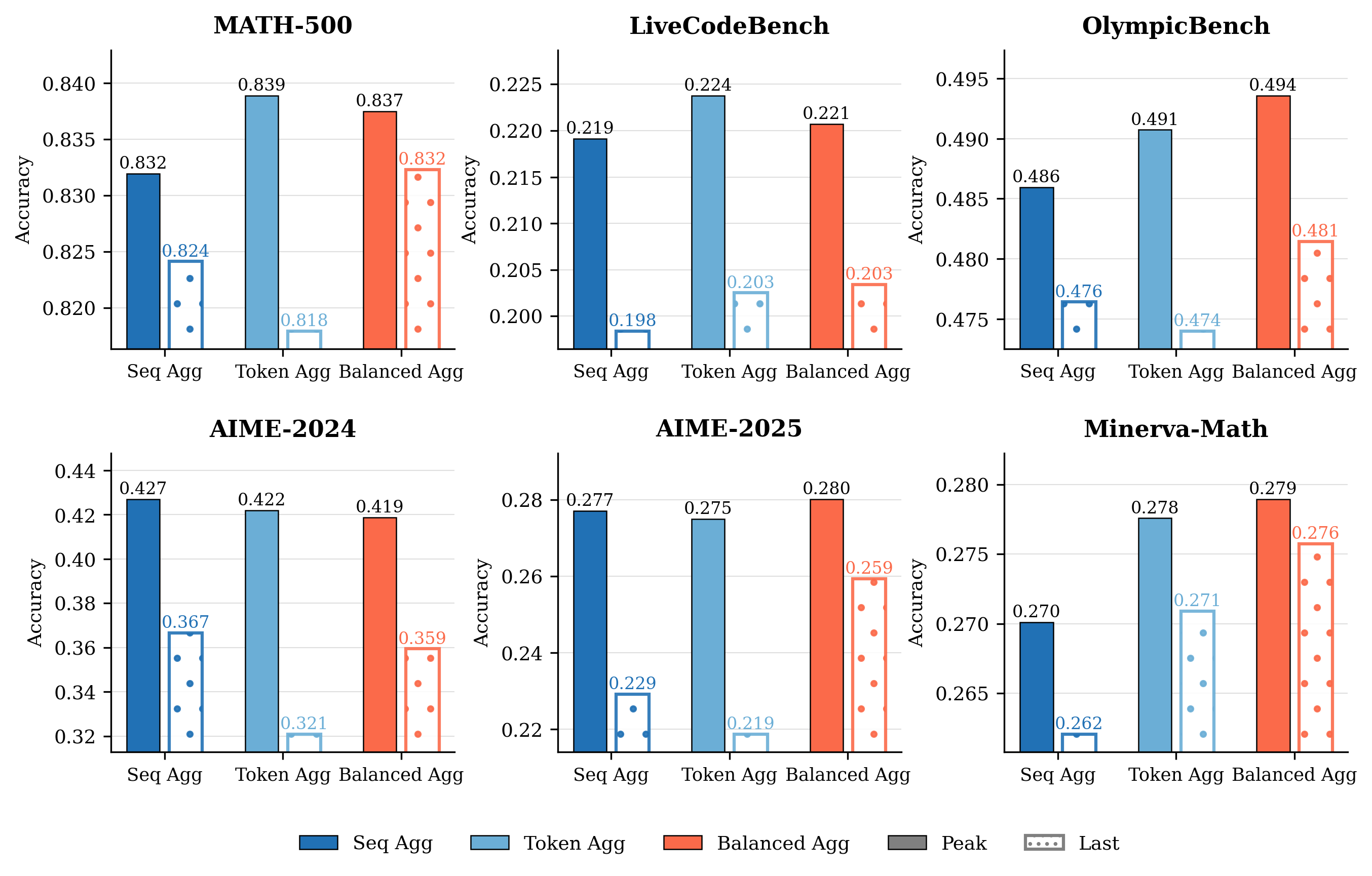}
\caption{Comparison of peak and last-step accuracy across evaluation benchmarks. For each benchmark, values are averaged over four training settings: Qwen2.5-Math-7B and Qwen3-1.7B, each trained on DAPO-17k and Polaris. Solid bars indicate peak performance, while dashed bars indicate last-step performance.}
\label{fig:peak_vs_last}
\end{figure*}

Figure~\ref{fig:peak_vs_last} compares peak and last-step accuracy on each benchmark, where each bar is averaged over four training settings: Qwen2.5-Math-7B and Qwen3-1.7B, each trained on DAPO-17k and Polaris. At peak performance, \textbf{token-agg} and \textbf{balanced-agg} are very close, and on most benchmarks both outperform \textbf{seq-agg}. The difference emerges at the final checkpoint: \textbf{token-agg} exhibits the largest peak-to-last drop on nearly all benchmarks, whereas \textbf{balanced-agg} preserves its gains much better and achieves the best or tied-best last-step result on five of the six benchmarks. The largest peak-to-last gaps appear on AIME-2024 and AIME-2025, likely due in part to the higher variance of their small evaluation sets. Overall, these results further indicate that \textbf{balanced-agg} delivers substantially stronger training stability than standard token aggregation while remaining highly competitive at peak performance.

\subsubsection{Policy-gradient loss dynamics and aggregation bias}
\begin{figure*}[tb]
\centering
\subfigure[Qwen2.5 on DAPO]{%
  \includegraphics[width=0.48\linewidth]{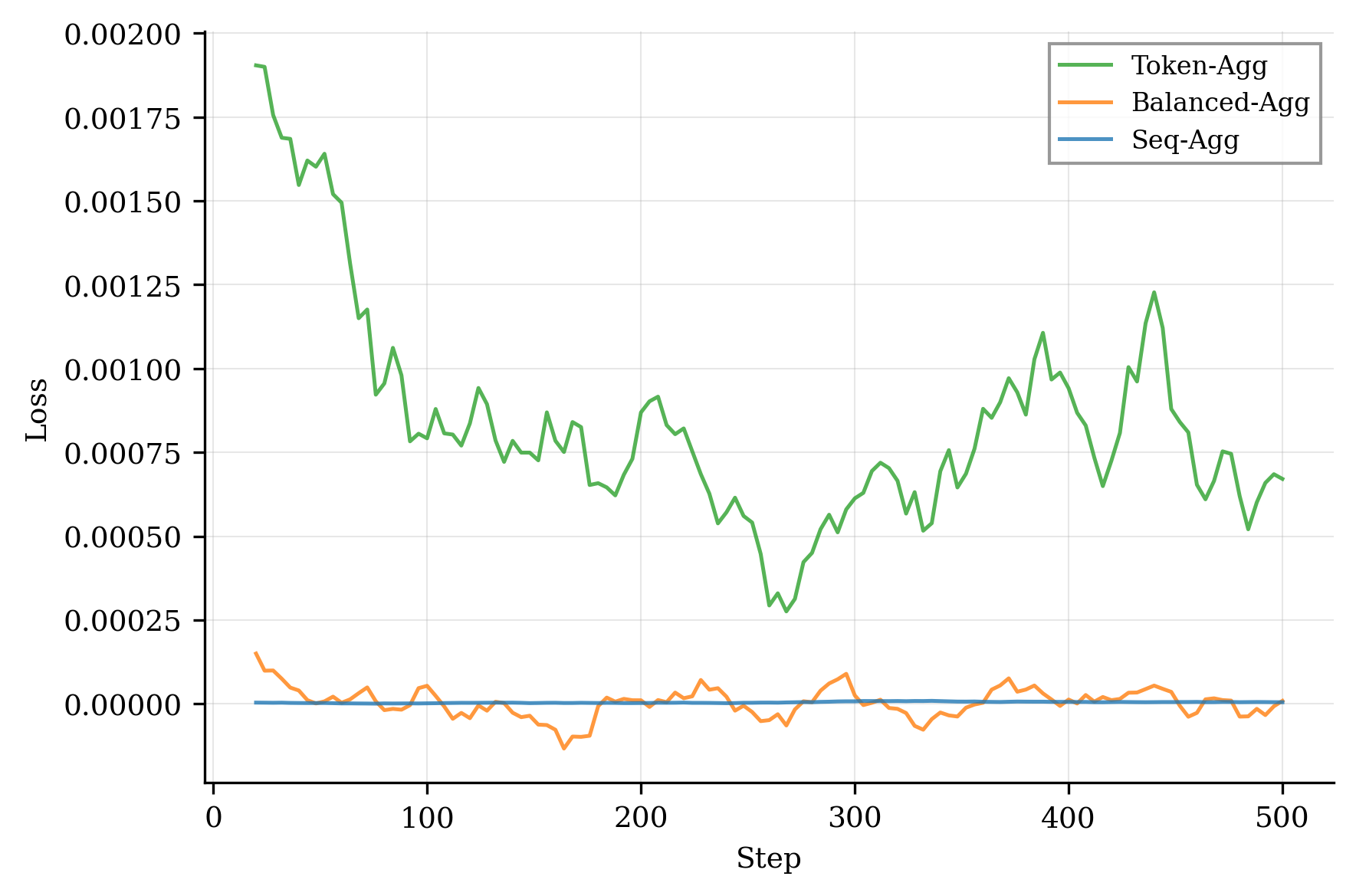}%
}\hfill
\subfigure[Qwen2.5 on Polaris]{%
  \includegraphics[width=0.48\linewidth]{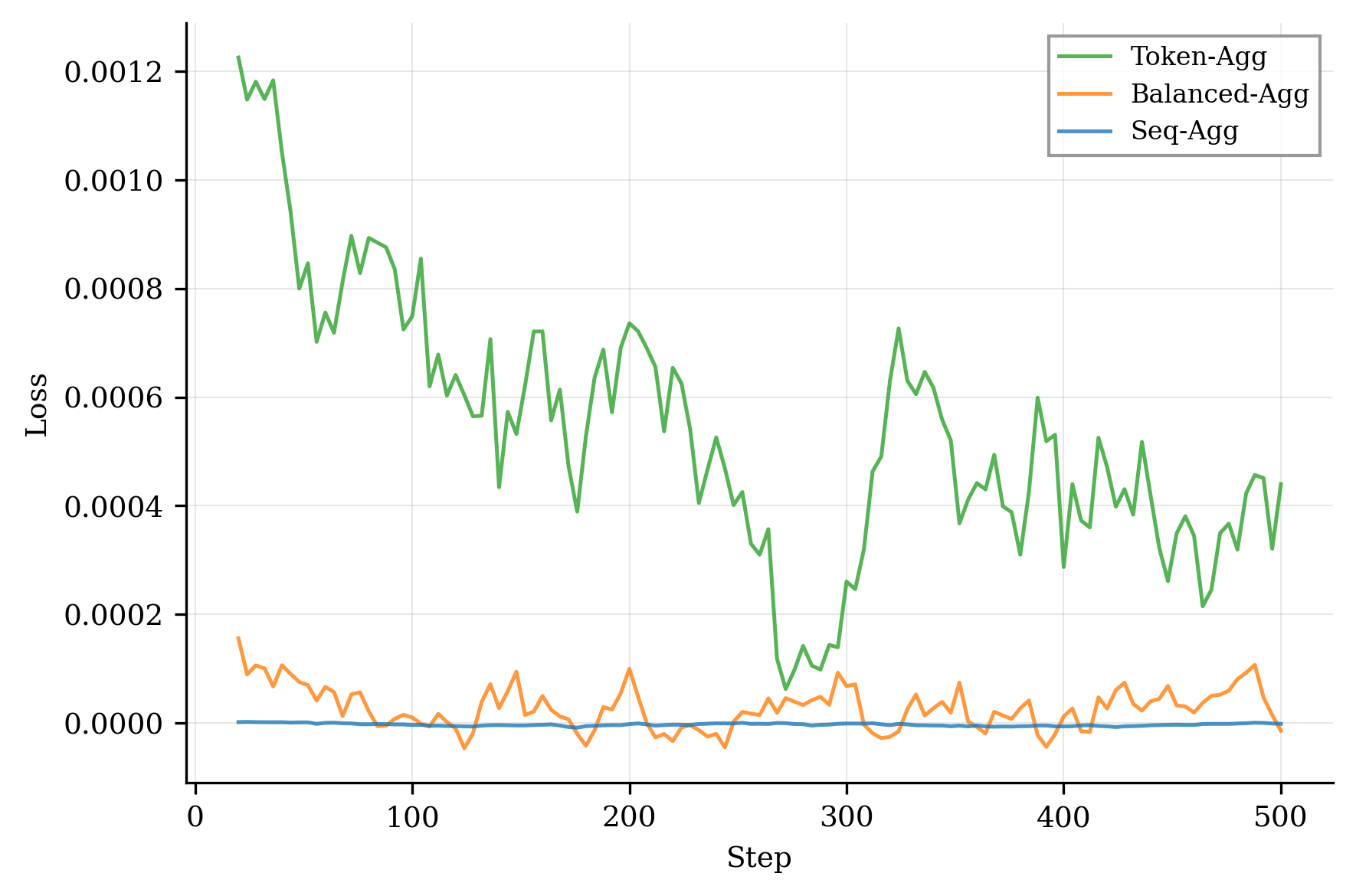}%
}
\caption{Training dynamics of the policy-gradient loss for Qwen2.5-Math-7B on DAPO-17k and Polaris. Triggered by the phenomenon that negative responses are systematically longer than positive responses, \textbf{token-agg} places disproportionate weight on negative samples, causing its loss to severely deviate from zero. In contrast, \textbf{seq-agg} and \textbf{balanced-agg} actively decouple this sign-length bias and remain much more stable.}
\label{fig:pg_loss_evidence}
\end{figure*}

To empirically validate our theoretical observations regarding aggregation bias, we analyze the training dynamics of the policy-gradient loss. Figure~\ref{fig:pg_loss_evidence} plots the evolution of this loss over the course of training. We observe a stark contrast between standard token-level aggregation and the sequence-balanced configurations: the loss for \textbf{token-agg} experiences massive drifts and oscillates consistently in a region far above zero. In contrast, both \textbf{seq-agg} and \textbf{balanced-agg} remain highly stable, with their loss trajectories oscillating tightly around zero. These empirical dynamics directly corroborate our earlier analysis of the positive-negative sign-length coupling bias. 
This behavior directly matches our earlier theory: when the positive-negative response-length gap is large, \textbf{token-agg} suffers from sign-length coupling, which skews the effective gradient balance and produces a persistent drift away from zero. By contrast, \textbf{seq-agg} and \textbf{balanced-agg} remove this coupling, so their loss trajectories remain much more stable and close to zero.

\begin{figure*}[tb]
\centering
\subfigure[DAPO-17k: response-length variation]{%
  \includegraphics[width=0.48\linewidth]{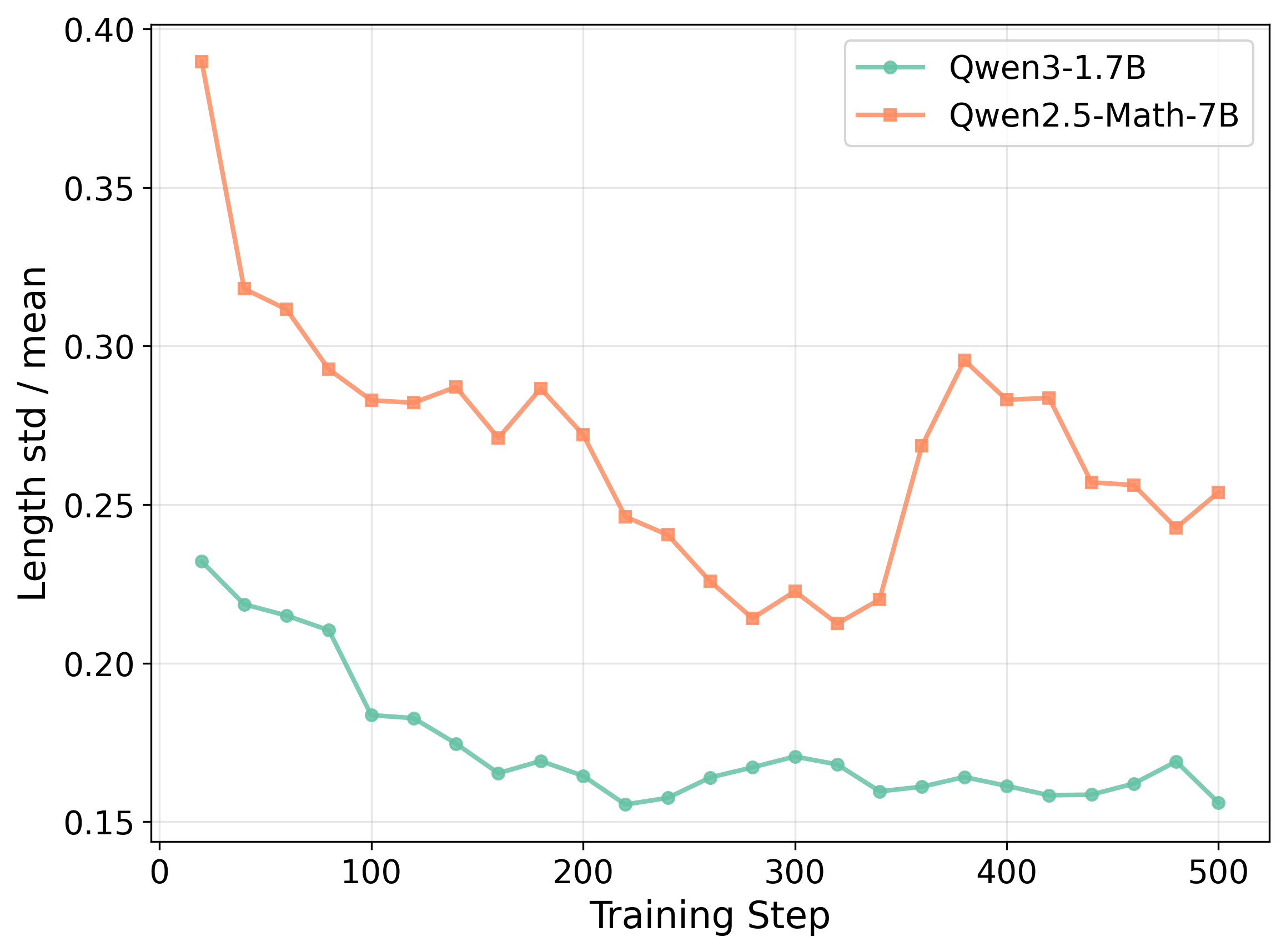}%
}\hfill
\subfigure[DAPO-17k: positive--negative length gap]{%
  \includegraphics[width=0.48\linewidth]{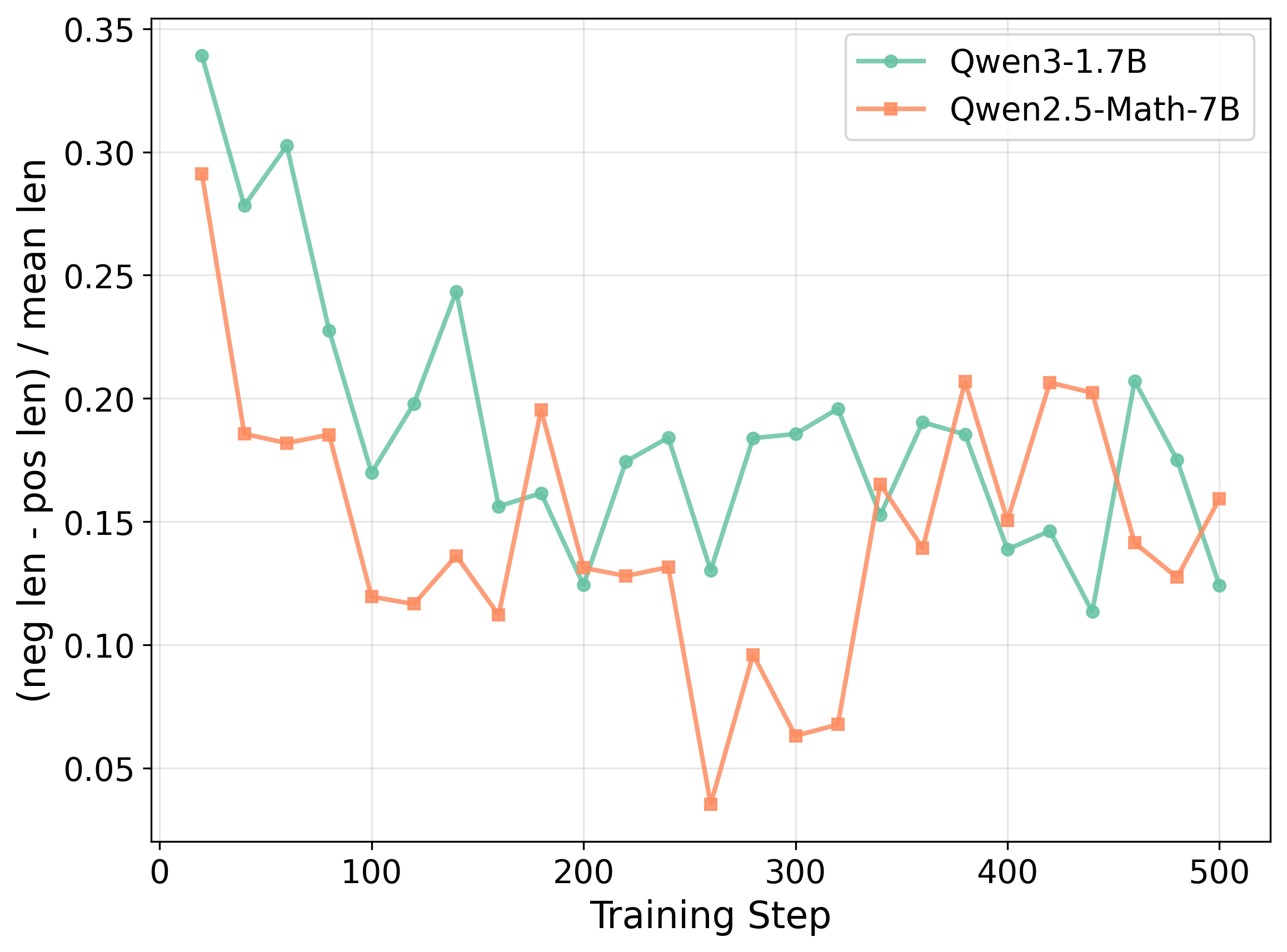}%
}

\vspace{0.6em}

\subfigure[Polaris: response-length variation]{%
  \includegraphics[width=0.48\linewidth]{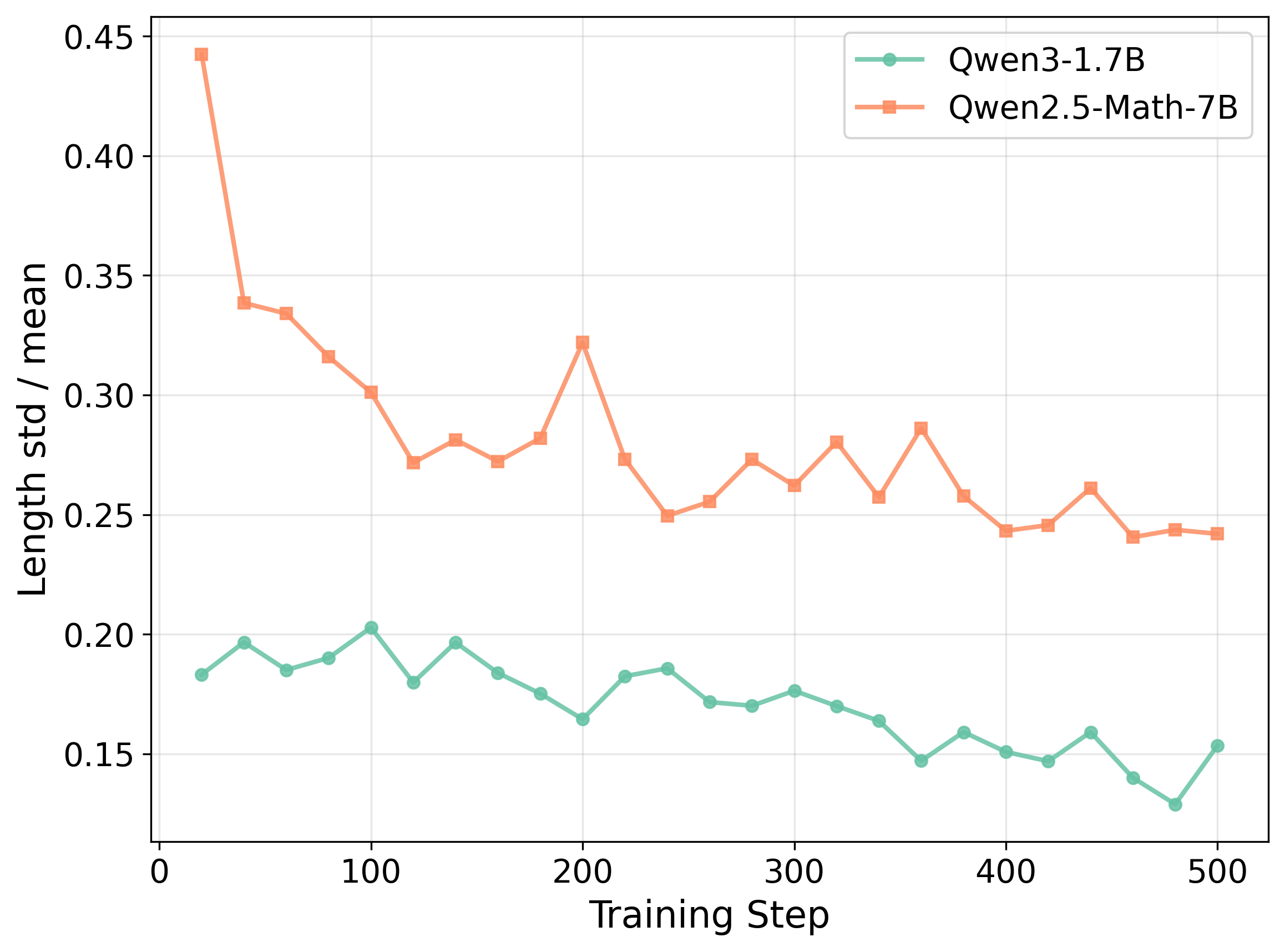}%
}\hfill
\subfigure[Polaris: positive--negative length gap]{%
  \includegraphics[width=0.48\linewidth]{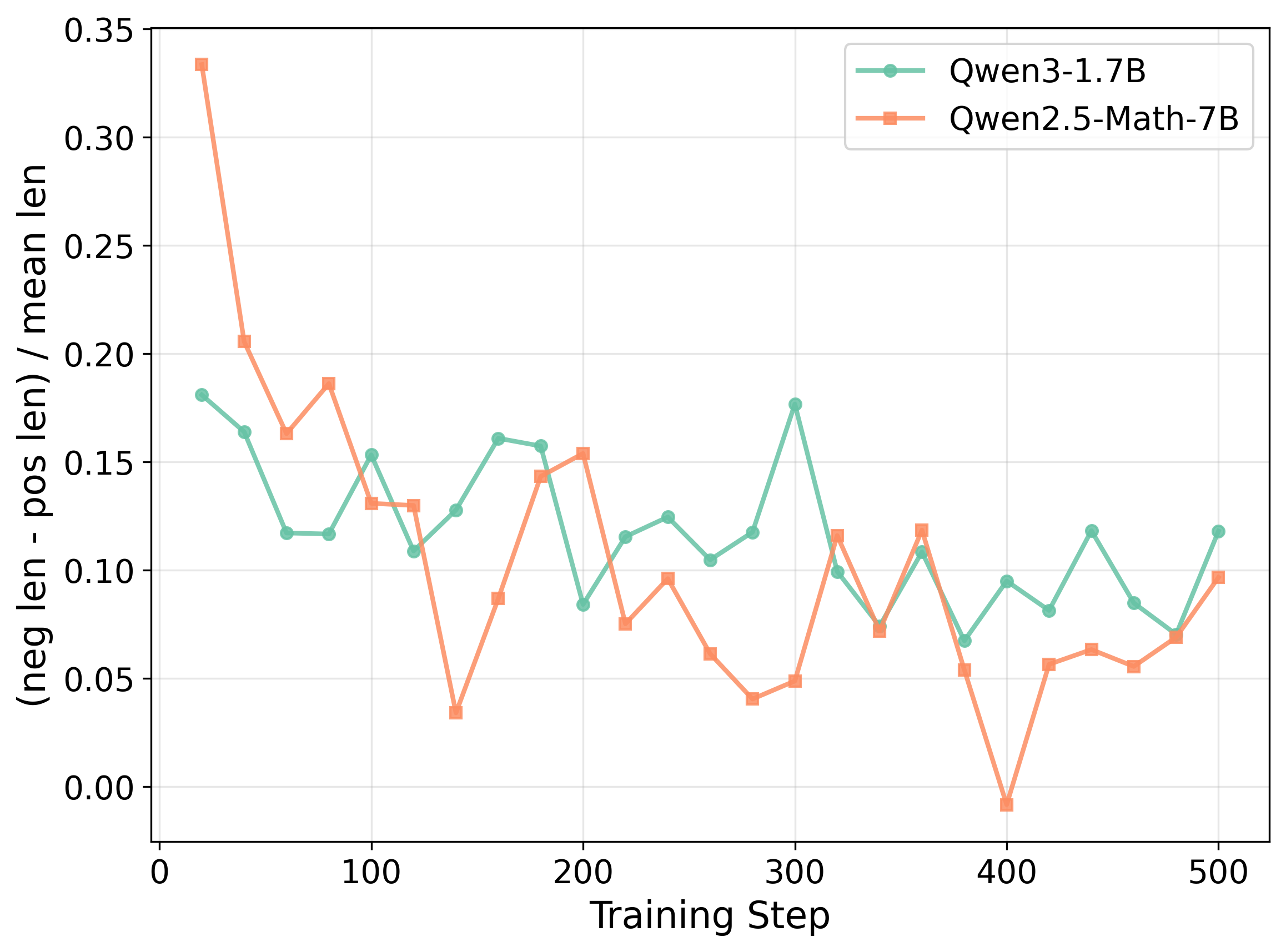}%
}
\caption{Length-distribution statistics related to the relative behavior of token and sequence aggregation. We report the response-length coefficient of variation and the positive--negative length gap over training on DAPO-17k and Polaris.}
\label{fig:length_analysis}
\end{figure*}

\subsubsection{Understanding the Model-Dependent Flip}
\label{sec:agg-theory-empirical}
\label{sec:why-token-agg-qwen25}

Our theory in Sections~\ref{sec:aggregation-rules} and \ref{sec:aggregation-bias} suggests a simple criterion for when each aggregation rule should be preferred. \textbf{Token-agg} should be more favorable when response-length variation is large but the positive--negative length gap is relatively mild, because in that regime the weakness of \textbf{seq-agg} is more pronounced than the weakness of \textbf{token-agg}. Conversely, \textbf{seq-agg} should be more favorable when response-length variation is small but the positive--negative length gap is large, because in that regime the sign-length coupling bias in \textbf{token-agg} becomes the dominant issue. In Section~\ref{sec:main-results}, we observe exactly such a model-dependent flip: \textbf{token-agg} performs better on Qwen2.5-Math-7B, whereas \textbf{seq-agg} performs better on Qwen3-1.7B. Figure~\ref{fig:length_analysis} tests this explanation by tracking over training the two quantities highlighted by the theory on DAPO-17k and Polaris.

\paragraph{When is Token-Agg Better than Seq-Agg?}
Qwen2.5-Math-7B falls into the regime favorable to \textbf{token-agg}. As shown in Figure~\ref{fig:length_analysis}, it exhibits substantially larger response-length variation than Qwen3-1.7B, while its positive--negative length gap is generally milder. This means the sequence-level equal-weighting effect in \textbf{seq-agg} is more harmful than the sign-length coupling bias in \textbf{token-agg}, which explains why \textbf{token-agg} can outperform \textbf{seq-agg} on Qwen2.5-Math-7B in Table~\ref{tab:main-results}.

\paragraph{When is Seq-Agg Better than Token-Agg?}
\label{sec:why-seq-agg-qwen3}
Qwen3-1.7B shows the opposite pattern. Its response-length variation is much smaller, so the weakness of \textbf{seq-agg} is reduced. Meanwhile, its positive--negative length gap is consistently larger, making the sign-length coupling bias in \textbf{token-agg} more severe. This is why \textbf{seq-agg} outperforms \textbf{token-agg} on Qwen3-1.7B in Table~\ref{tab:main-results}. The same comparison also helps explain why \textbf{balanced-agg} is robust across both regimes: it mitigates the positive--negative coupling that hurts \textbf{token-agg} without paying the full cost of sequence-level equal weighting.

\section{Conclusion}

We studied how aggregation in GRPO-style RLVR shapes optimization behavior. Token aggregation introduces sign-length coupling, while sequence aggregation avoids this coupling at the cost of strong per-sequence equal weighting. Balanced Aggregation (BA) provides a simple alternative that preserves sign balance while retaining token-level averaging within each sign group. Across models, datasets, and benchmarks, BA delivers more robust training and strong final performance. Overall, our results show that aggregation is a first-class design choice in GRPO-style RLVR, and that balancing inter-sign weighting without discarding within-sign token information leads to more stable optimization.

\clearpage
\bibliography{references}
\bibliographystyle{plainnat}

\appendix
\section*{Appendix A: Why Use Sequence-Count Weights in BA?}
\label{app:ba-weights}

Here we briefly justify the choice of weights \(k/G\) and \((G-k)/G\) in BA. Recall that under the binary-reward GRPO setting, the normalized advantages are
\begin{equation}
\hat A_i =
\sqrt{\frac{G-k}{k}}
\quad \text{for } i \in S_+,
\qquad
\hat A_i =
-\sqrt{\frac{k}{G-k}}
\quad \text{for } i \in S_-.
\end{equation}
Substituting these values into the BA objective gives
\begin{equation}
\mathcal J_{\mathrm{BA}}
=
\frac{k}{G}
\left(
\sqrt{\frac{G-k}{k}}\,\bar\delta_+^{\mathrm{BA}}
\right)
+
\frac{G-k}{G}
\left(
-\sqrt{\frac{k}{G-k}}\,\bar\delta_-^{\mathrm{BA}}
\right).
\end{equation}
Rearranging yields
\begin{equation}
\mathcal J_{\mathrm{BA}}
\propto
\frac{\sqrt{k(G-k)}}{G}
\left(
\bar\delta_+^{\mathrm{BA}} - \bar\delta_-^{\mathrm{BA}}
\right),
\end{equation}
which is exactly the same inter-sign prefactor as sequence aggregation in Eq.~\eqref{eq:seq-balanced-form}. Therefore, the sequence-count weights \(k/G\) and \((G-k)/G\) are not arbitrary: they are precisely what ensures that BA preserves the same positive-negative balancing principle as sequence aggregation, while still using token-level averaging within each sign group.

\section*{Appendix B: Extension to Non-Binary Rewards}
\label{app:nonbinary}
Our main presentation of Balanced Aggregation (BA) focuses on the standard binary-reward GRPO setting, where the normalized advantages are constant within each sign subset. In that case, the positive and negative groups can be characterized solely by their sequence counts, which leads to the simple sequence-count weights \(k/G\) and \((G-k)/G\).

For general real-valued rewards, however, the normalized advantages \(\hat A_i\) are no longer constant even within the same sign subset. As a result, sequence counts alone are no longer sufficient to characterize the relative contribution of the positive and negative subsets. In particular, if we write the token-level PPO contribution as
\[
\phi_{i,t}(\theta)=\hat A_i\,\delta_{i,t}(\theta),
\]
then the total contribution of each sign subset depends not only on the number of responses and their lengths, but also on the magnitudes of their advantages.

To generalize BA to this setting, we first define the positive and negative subsets
\[
S_+ = \{i \mid \hat A_i > 0\},
\qquad
S_- = \{i \mid \hat A_i < 0\}.
\]
We then define the corresponding sign-wise advantage masses
\[
M_+ = \sum_{i \in S_+} \hat A_i,
\qquad
M_- = \sum_{i \in S_-} (-\hat A_i),
\]
and the sign-wise advantage-weighted token masses
\[
Z_+ = \sum_{i \in S_+} \hat A_i T_i,
\qquad
Z_- = \sum_{i \in S_-} (-\hat A_i) T_i.
\]

Using these quantities, we define the generalized BA objective as
\[
\mathcal J_{\mathrm{BA\text{-}gen}}(\theta)
=
\mathbb E\left[
\frac{M_+}{G}
\cdot
\frac{1}{Z_+}
\sum_{i \in S_+}\sum_{t=1}^{T_i}\phi_{i,t}(\theta)
+
\frac{M_-}{G}
\cdot
\frac{1}{Z_-}
\sum_{i \in S_-}\sum_{t=1}^{T_i}\phi_{i,t}(\theta)
\right].
\]

Since \(\phi_{i,t}(\theta)\) is positive-advantage-weighted on \(S_+\) and negative-advantage-weighted on \(S_-\), this construction preserves the original policy-gradient weighting induced by \(\hat A_i\), while normalizing the positive and negative subsets separately. Equivalently, substituting \(\phi_{i,t}(\theta)=\hat A_i\delta_{i,t}(\theta)\) gives
\[
\mathcal J_{\mathrm{BA\text{-}gen}}(\theta)
=
\mathbb E\left[
\frac{M_+}{G}\,\bar\delta_+^{\mathrm{gen}}
-
\frac{M_-}{G}\,\bar\delta_-^{\mathrm{gen}}
\right],
\]
where
\[
\bar\delta_+^{\mathrm{gen}}
=
\frac{1}{Z_+}
\sum_{i \in S_+}\hat A_i \sum_{t=1}^{T_i}\delta_{i,t}(\theta),
\qquad
\bar\delta_-^{\mathrm{gen}}
=
\frac{1}{Z_-}
\sum_{i \in S_-}(-\hat A_i)\sum_{t=1}^{T_i}\delta_{i,t}(\theta).
\]

Therefore, the generalized BA retains the same core principle as the binary version: positive and negative samples are first normalized within their own sign subsets and are then recombined in a sign-balanced manner. The key difference is that, in the non-binary case, sign balance is determined by advantage mass rather than sequence count.

Moreover, under the usual group-normalization condition
\[
\sum_{i=1}^G \hat A_i = 0,
\]
we have
\[
M_+ = M_- = \frac{1}{2}\sum_{i=1}^G |\hat A_i|.
\]
Hence, the positive and negative subsets remain symmetric at the inter-sign level, so the generalized BA still removes the cross-sign sign-length coupling induced by standard token aggregation.

Finally, under binary rewards, \(\hat A_i\) is constant within each sign subset. In that case,
\[
M_+ = k a_+,\qquad M_- = (G-k)a_-,
\qquad
Z_+ = a_+ N_+,\qquad Z_- = a_- N_-,
\]
for some constants \(a_+>0\) and \(a_->0\). Substituting these expressions into \(\mathcal J_{\mathrm{BA\text{-}gen}}\) recovers exactly the original BA objective defined in the main text. Therefore, the generalized formulation is a strict extension of BA rather than a different objective.

\end{document}